\documentclass[journal]{IEEEtran}
\ifCLASSINFOpdf
\else
\fi

\usepackage{cite}
\usepackage{graphicx}
\usepackage{multicol}
\usepackage{subfig}
\usepackage{amsmath}
\usepackage{amsfonts}
\usepackage{romannum}
\usepackage{caption}
\usepackage{tabularx}
\usepackage{tabulary}
\usepackage{array}
\usepackage{tabu}
\usepackage{xcolor}

\DeclareMathOperator*{\argmax}{arg\,max}  % in your preamble
\begin{document}

\title{Liver Segmentation in Abdominal CT Images \\via Auto-Context Neural Network \\and Self-Supervised Contour Attention}

\author{Minyoung Chung, Jingyu Lee, Jeongjin Lee$^{\ast}$, and Yeong-Gil Shin%
\thanks{\textit{Asterisk indicates corresponding author.}}%
\thanks{M. Chung, J. Lee, and Y.-G. Shin are with the Department of Computer Science and Engineering, Seoul National University, Korea (e-mail: chungmy@snu.ac.kr).}%
% \thanks{J. Lee is with the Department of Computer Science and Engineering, Seoul National University, Korea (e-mail: leejingyu@cglab.snu.ac.kr).}%
\thanks{*J. Lee is with the Department of Computer Science and Engineering, Soong-sil University, Korea (e-mail: profjjlee@naver.com).}%
% \thanks{J. W. Chung is with the Department of Radiology, Seoul National University Hospital, Korea (e-mail: chungjw@snu.ac.kr).}}

% \thanks{This work was partly supported by Institute for Information \& communications Technology Promotion(IITP) grant funded by the Korea government(MSIT) (No.2017-0-018715, Development of AR-based Surgery Toolkit and Applications). And this research was supported by the Basic Science Research Program through the National Research Foundation of Korea (NRF) funded by the Ministry of Science, ICT and Future Planning (No. 2017R1D1A1B03034484). And this work was partly supported by Institute for Information \& Communications Technology Promotion(IITP) grant funded by the Korean government(MSIP). [R0118-16-1003, Authoring Platform Technology for Next-Generation Plenoptic Contents].}
}

% The paper headers
% \markboth{IEEE TRANSACTIONS ON MEDICAL IMAGING}%
% {Shell \MakeLowercase{\textit{et al.}}: Bare Demo of IEEEtran.cls for IEEE Journals}

\maketitle

\begin{abstract}
Accurate image segmentation of the liver is a challenging problem owing to its large shape variability and unclear boundaries. Although the applications of fully convolutional neural networks (CNNs) have shown groundbreaking results, limited studies have focused on the performance of generalization. In this study, we introduce a CNN for liver segmentation on abdominal computed tomography (CT) images that shows high generalization performance and accuracy. To improve the generalization performance, we initially propose an auto-context algorithm in a single CNN. The proposed auto-context neural network exploits an effective high-level residual estimation to obtain the shape prior. Identical dual paths are effectively trained to represent mutual complementary features for an accurate posterior analysis of a liver. Further, we extend our network by employing a self-supervised contour scheme. We trained sparse contour features by penalizing the ground-truth contour to focus more contour attentions on the failures. The experimental results show that the proposed network results in better accuracy when compared to the state-of-the-art networks by reducing 10.31\% of the Hausdorff distance. We used 180 abdominal CT images for training and validation. Two-fold cross-validation is presented for a comparison with the state-of-the-art neural networks. Novel multiple N-fold cross-validations are conducted to verify the performance of generalization. The proposed network showed the best generalization performance among the networks. Additionally, we present a series of ablation experiments that comprehensively support the importance of the underlying concepts.
\end{abstract}

% Note that keywords are not normally used for peerreview papers.
\begin{IEEEkeywords}
Auto-context neural network, contour attention network, high-level residual shape prior estimation, liver segmentation.
\end{IEEEkeywords}

\IEEEpeerreviewmaketitle

\section{Introduction}

% brief domain introduction.
\IEEEPARstart{M}{edical} image segmentation is an essential prerequisite for clinical applications of a computer-aided diagnosis system, such as volume measurement, treatment planning, and further virtual or augmented surgeries \cite{van2011computer, howe1999robotics}. Among the organs, the liver is a highly in demand as liver diseases are among the primary increasing causes of death worldwide \cite{bray2018global}. For accurate surgical planning, such as liver transplantation and resection, volumetric information of the liver is critically required. However, manual or semi-automatic image segmentation of the liver is an impractical task owing to its large shape variability and unclear boundaries. Unlike other organs, ambiguous boundaries with the heart, stomach, pancreas, and the occurrence of fat result in difficulty in the image segmentation of the liver. Furthermore, manual segmentation is error-prone, which implies that there exists a severe inter- and intra-observer variability in the results.\par

% classical liver segmentation methods.
A vast body of literature on automatic liver segmentation has been previously presented. Many classical methods, before the era of deep learning, employed image- or shape-based approaches \cite{suzuki2010computer, lee2007efficient, zhang2010automatic, ling2008hierarchical, heimann2009comparison}. Among them, an active contour model (ACM) was a popular approach, which regards the segmentation task as a contour delineation \cite{suzuki2010computer, lee2007efficient}. The ACM approach attempts to design an objective energy functional that drives the contour to propagate toward the target object by numerical optimization techniques \cite{caselles1995geodesic, chan2001active, osher1988fronts}. However, the stopping criteria of ACM primarily rely on the local intensity distribution, which easily breaks down owing to the large variance in foreground intensity distribution and unclear boundaries of the liver. Conversely, shape-based methods, such as an active/statistical shape model, were developed to overcome such difficulties \cite{ling2008hierarchical, zhang2010automatic, heimann2007statistical, wimmer2009generic, van2007automatic}. Shape-based methods are regarded as more successful approaches than simple intensity-based methods owing to the embedded shape priors. However, the shape-based methods also suffer from limited prior information (i.e., lack of liver database) as it is difficult to embed all inter-patient organ shapes. Moreover, fine registration is still challenging owing to irregular boundaries.\par
% Thus, the number of data that are used to train statistical models directly affects the performance of the segmentation.\par

% deep learning (medical) segmentation methods (common limitations).
With the advent of deep learning, convolutional neural networks (CNNs) have been showing promising results over the conventional methods for the medical image segmentation task \cite{ronneberger2015u, cciccek20163d, milletari2016v, chen2017voxresnet, chen2017dcan, kamnitsas2017efficient, havaei2017brain, dou20173d, oktay2018anatomically, gibson2018automatic, schlemper2019attention}. However, the performance of generalization was not addressed, which is the most important feature in the actual deployment of CNNs for medical image segmentation tasks. Many studies were conducted to obtain a high generalization performance of neural networks, such as weight decay, drop out \cite{srivastava2014dropout}, transfer learning \cite{pan2009survey}, data augmentation \cite{devries2017improved}, domain adaptation \cite{cai2019unsupervised, long2015learning}, and regularization of loss functions \cite{pereyra2017regularizing}. However, these global, systematic techniques demonstrate limitations in adapting to other fields that have severe data deficiency and intrinsic class imbalance (e.g., rare cases of anomalies and phases in medical images). Thus, a domain-specific generalization technique is highly required, especially in the field of medical image analysis. Additionally, it is worth knowing that the image segmentation problem can be resolved by delineating the accurate boundaries of an object in the image, such as in ACM approaches \cite{caselles1995geodesic, chan2001active, osher1988fronts, suzuki2010computer, lee2007efficient}. However, research focusing on the implantation of a contour scheme to modern end-to-end CNN frameworks is significantly limited. Oppose to a previous study \cite{chen2017dcan}, it is difficult to simply supervise a CNN to delineate the full ground-truth contour in a multi-task framework because many ambiguous boundaries exist on the liver border. In this study, we propose a novel CNN architecture to address the aforementioned issues regarding generalization performance and contour scheme implantation.\par

\begin{figure}[t]
    \centering
    \includegraphics[width=\linewidth]{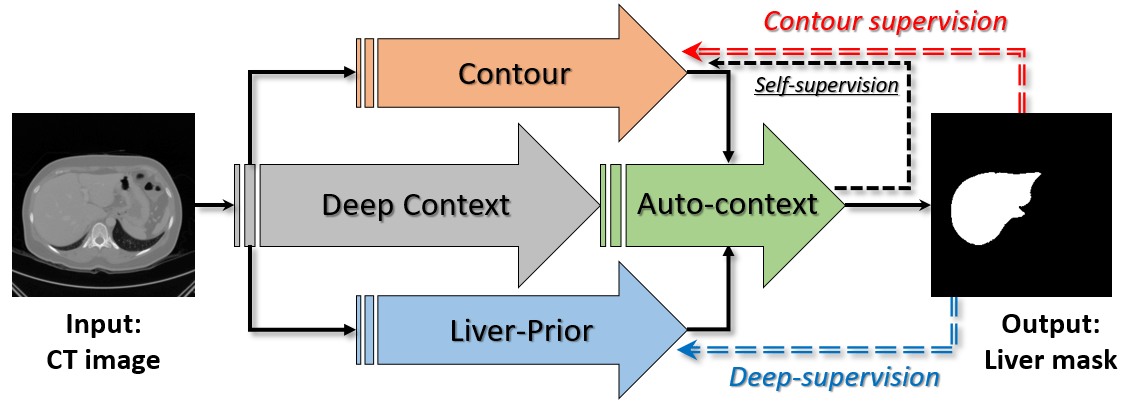}
    \caption{Overall architecture of the proposed neural network.}
    \label{fig:brief}
\end{figure}

% brief introduction of the proposed network architecture.
The base architecture of our network is an auto-context algorithm \cite{tu2010auto}. We employed the auto-context algorithm \cite{tu2010auto} to a single neural network by using a liver-prior branch (Fig. \ref{fig:brief}). The liver-prior branch is deeply supervised to generate the probability of a liver foreground. The prior is then fused with deep contexts for the final auto-context layers. In addition to the auto-context structure, we added another branch, which is also deeply supervised to delineate the contour of a liver. Instead of training the explicit ground-truth contour, we trained sparse contours by a self-supervising method that acts as an implicit contour attention. The self-supervision is obtained by the final prediction of the network, which penalizes the ground-truth contour image based on confidence. The primary underlying principle of the proposed architecture is that the accurate segmentation of a liver can be achieved by a robust shape prior and an accurate contour delineation. This work is an extension of our previous work on a contour embedded network (CENet) \cite{chung2018deeply}, which employed the self-supervised contour embedding. The proposed network in this study automated the previous self-supervision of the contour by removing categorical classification loss that was formed by a heuristic threshold value \cite{chung2018deeply}. We referred to our proposed network as automated auto-context CENet (AutoCENet). The network also reduced a large number of parameters based on the compact formulation of the auto-context algorithm.\par

% overview of the remainders.
The remainder of this paper is organized as follows. In Section \Romannum{2}, several CNN models, auto-context algorithms, and contour embedding mechanisms are reviewed. The proposed method is described in Section \Romannum{3}. The experimental results, discussion, and conclusion are presented in Sections \Romannum{4}, \Romannum{5}, and \Romannum{6}, respectively.

\section{Related Works}
\subsection{CNNs for Medical Image Segmentation}
Since a fully convolutional network (FCN) was introduced, several CNN architectures have been developed for medical image segmentation tasks. To extract 3D anatomical contexts, a 3D U-net \cite{cciccek20163d} was presented by replacing all the 2D convolutional operators in the original U-net with their 3D counterparts. The U-net architecture employs contracting and expanding paths together with skip connections, which combines both low- and high-level features \cite{ronneberger2015u}. In \cite{milletari2016v}, a full 3D CNN-based U-net-like architecture was presented to segment volumetric medical images using dice coefficient that tackles the class imbalance problem. The dice loss presented in \cite{milletari2016v} intrinsically overcame the class imbalance problem by avoiding a strong bias toward background learning. A deep contour-aware network was developed to depict clear contours by designing a multi-task framework \cite{chen2017dcan}. A voxelwise residual network (VoxResNet) \cite{chen2017voxresnet} performed brain tissue segmentation by employing voxelwise residual connections. Additionally, the authors employed an auto-context algorithm to further refine the voxelwise prediction results \cite{chen2017voxresnet}. A deep supervision mechanism \cite{lee2015deeply} was employed to supervise multiple intermediate layers, which enhanced the discriminability of the low-level features \cite{dou20173d}. The authors argued that when more discriminable low-level features are extracted, a more discriminative final classification can be obtained, which results in the improvement of the generalization performance \cite{dou20173d}. A densely connected convolutional architecture \cite{huang2017densely} was employed by designing a similar architecture as a V-net for the task of multiorgan segmentation \cite{gibson2018automatic}. The singularity of the network was the introduction of a trainable grid that learns the shape prior \cite{gibson2018automatic}. More recently, the attention mechanism was successfully employed in the 3D U-net architecture to boost the performance of the network in \cite{schlemper2019attention}. The authors hierarchically applied the attention gate module to disambiguate task-irrelevant feature contexts in the intermediate layers (AGU-net) \cite{schlemper2019attention}.

\subsection{Auto-Context Algorithm}
The auto-context mechanism fuses implicit shape information and low-level appearance features to perform image segmentation \cite{tu2010auto}. The posterior distribution of the given segmentation problem is learned with marginal distribution (i.e., classified probability map), which is further combined to learn the final classifiers. The posterior marginal distribution is learned through image patches by calculating the following distribution \cite{tu2010auto}:

\begin{equation}
    p(y_i|\bold{x})=\int p(y_i, \bold{y}_{-i}|\bold{x})d\bold{y}_{-i},
\label{eq:marginal_dist}
\end{equation}

\noindent
where $\bold{x}$, $\bold{y}$ present a given image and ground-truth label vector, respectively, and $\bold{y}_{-i}$ is a marginal set, $\{\bold{y}-y_i\}$. We have omitted patch representation for simplicity. Traditional feature extractors (e.g., Haar \cite{viola2004robust}, histogram of oriented gradients \cite{dalal2005histograms}) and classifiers (e.g., probabilistic boosting tree \cite{tu2005probabilistic}) were used for patch-wise prediction for the calculation (\ref{eq:marginal_dist}). The algorithm iteratively solves the posterior probability with the previous marginal distribution:

\begin{equation}
    p^{(t)}(y_i|\bold{x}, \bold{\Tilde{p}}^{(t-1)}) \longrightarrow p(y_i|\bold{x}),
\label{eq:posteior}
\end{equation}

\noindent
where $\bold{\Tilde{p}}^{(t-1)}$ is a posterior marginal for each pixel i learned according to (\ref{eq:marginal_dist}). It was proven by the authors that the algorithm asymptotically converges to $p(y_i|\bold{x})$ with a discrete, iterative process. In contrast to the original paper \cite{tu2010auto}, we used the term \textit{``context"} in this paper as a feature used in the second classifier (i.e., not shape information).

\begin{figure*}[t!]
    \centering
    \includegraphics[width=\linewidth]{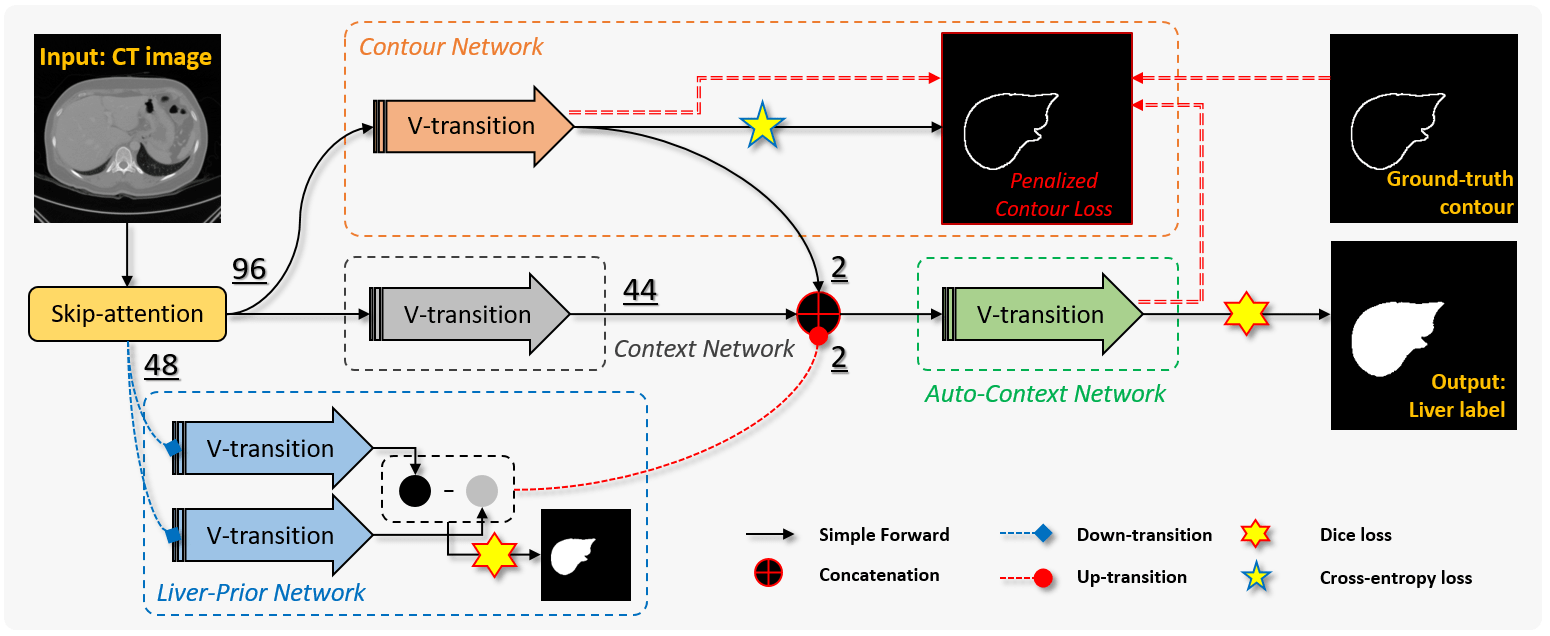}
    \caption{Proposed 3D network architecture. Stacked V-transitions form a base module with multiple skip connections. The red (i.e., circled arrow) and blue (i.e., squared) arrows indicate up- and down-transition layers, respectively. The red and blue dotted boxes represent the contour and liver-prior transitions, respectively. The two transitions are deeply supervised by the penalized contour and ground-truth liver foreground. The penalized contour loss is formulated by employing the final output, output of the contour network, and the ground-truth contour labels (double lines with red arrows). All the images are displayed in 2D for simplicity. This image is best viewed in color.}
    \label{fig:network}
\end{figure*}

\subsection{Self-Supervised Contour Embedding}
The contour features were successfully embedded in the network in \cite{chung2018deeply}. The authors deeply supervised the contour extraction layer by a dynamic modification of the ground-truth contour for each iteration, as presented below:
\begin{equation}
    \Tilde{\Gamma_c}=\Gamma_c \otimes (\bold{\Tilde{y}}_p),
\label{eq:closs1}
\end{equation}
where \(\otimes\) is an element-wise multiplication operator, $\Gamma_c$ is the ground-truth contour image, and \(\bold{\Tilde{y}}_p\) is a binary image with respect to the threshold value \(p\):
\newline
\begin{equation}
    \bold{\Tilde{y}}_p(x)=
    \begin{cases}
    1, & \text{if}\ \bold{\Tilde{y}}(x)<p,\\
    0, & \text{otherwise},
    \end{cases}
\label{eq:closs2}
\end{equation}
where $\bold{\Tilde{y}}$ is the output probability prediction of the proposed network for a given iteration. That is, the ground-truth contours were automatically erased if the network successfully delineated the corresponding labels at the output. A manually self-supervised contour embedding mechanism was established by explicit attention to the misclassified contour region. The training of the contour feature was performed by a cross-entropy classification loss for each voxel.

% \subsection{Attention Mechanism and Penalizing Confident Outputs}
% The attention mechanism \cite{wang2017residual, hu2018squeeze, woo2018cbam. jetley2018learn} and penalizing the confident output distribution [] are similar in the perspective of internally weighting the neural network to boost the accuracy. The attention method is applied to the intermediate layers that weighs the feature maps by either channel- or spatial-wise manner \cite{hu2018squeeze, woo2018cbam}. On the other hand, penalizing the output method attempts to modify the loss function to regularize the network []. There are many studies that relate the self-attention neural networks to the medical image segmentation tasks []. \textcolor{blue}{However, unlike channel-wise attentions [], spatial-attention is not suitable because it is difficult to define specific spatial locations to focus on [\textit{compare}].} In this work, we employed an implicit channel-wise attention mechanism to the various non-linear modules of the proposed network. We also applied the contour attention based on the penalizing output method that is explicitly modeled. The contour self-supervision, proposed in \cite{chung2018deeply}, is similar to penalizing the confident output distributions, however, the proposed mechanism was manually designed to manipulate the important regions of contours. The previous exclusive method (i.e., true or false in the contour classification) is furthermore ineffective in the perspective of instability during training.

\begin{figure}[t]
    \centering
    \subfloat[Skip-attention layer (common feature extraction module). The intermediate features are skip-connected by concatenation. A trainable channel-wise attention vector is employed for the final output features. The number of output features is 96.]{\includegraphics[width=\linewidth]{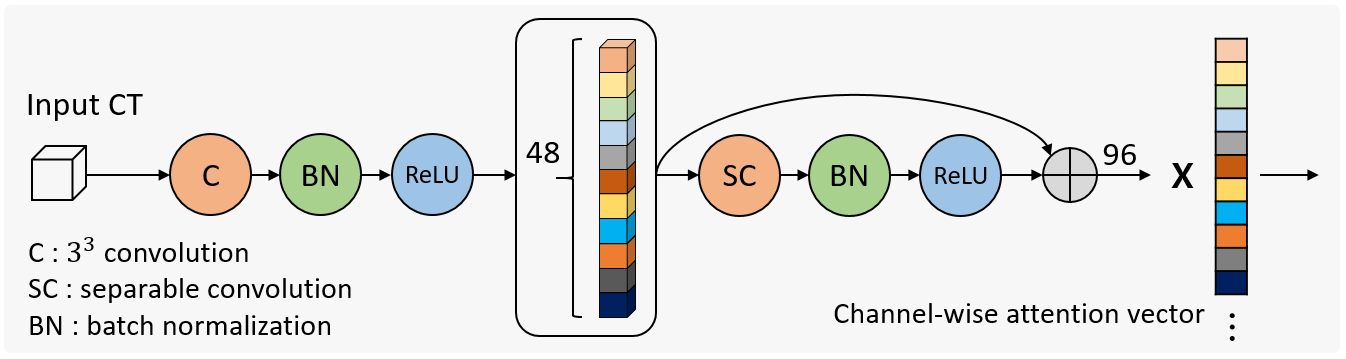}%
    \label{fig:layers_skip}}
    \vfil
    \subfloat[V-transition layer. A structured non-linearity module (i.e., SC-BN-ReLU) composed of a series of separable convolutions (SCs), batch normalizations (BNs), and rectified linear units (ReLUs). A multiscaled feature analysis is applied by down-transition through $2^3$ convolutions with stride = 2 and up-transition through $2^3$ transposed convolutions with stride = 2. A skip-connection and channel-wise attention vector are employed in the lower resolution similar to the skip-attention block. The final output is obtained by a $1^3$ convolution applied to the concatenated features.]{\includegraphics[width=\linewidth]{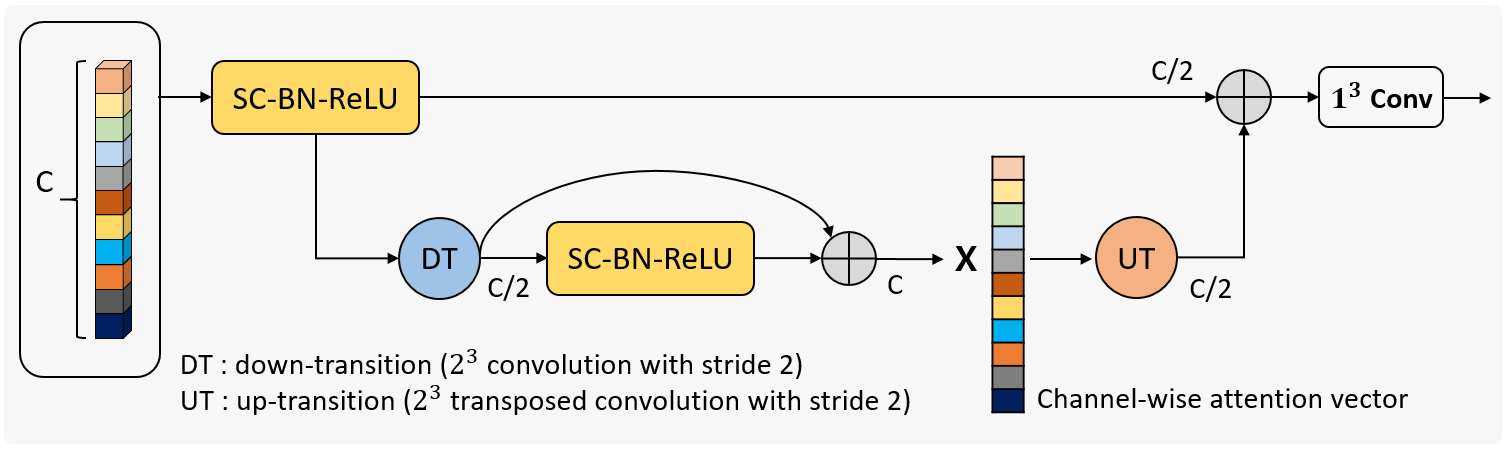}%
    \label{fig:layers_v}}
    \vfil
    \subfloat[Depth-wise separable convolutions. The input channels are separated by groups and are convolved separately. The final output is a concatenation of all groups. The number of groups is four in the proposed network.]{\includegraphics[width=\linewidth]{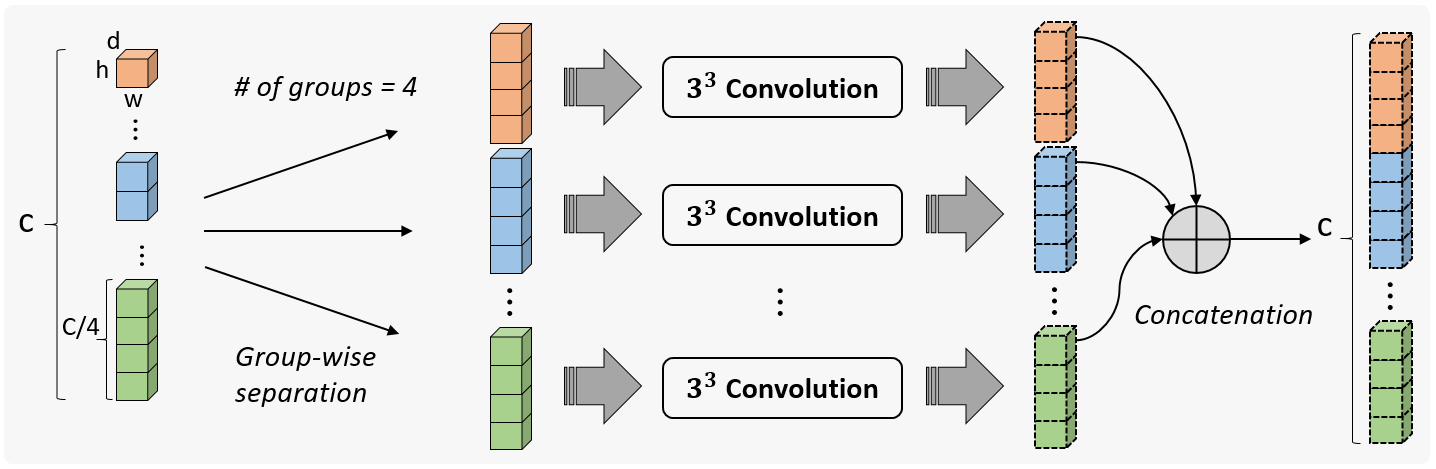}%
    \label{fig:layers_sep}}
    
    \caption{Proposed non-linear layers employed in the proposed network: (a) skip-attention, (b) V-transition layers, and (c) depth-wise separable convolutions.}
    \label{fig:layers}
\end{figure}

\section{Methodology}
% General overview.
The proposed network architecture is composed of three primary branches: liver-prior, context, and contour, i.e., the blue, gray, and orange dotted boxes in Fig. \ref{fig:network}, respectively. The liver-prior network is deeply supervised to estimate the ground-truth liver in a lower resolution. The trained posterior (i.e., the output of the liver-prior network) is used as a prior in the remaining auto-context network. Deep features that are trained by the context network are concatenated to the prior for the final auto-context fusion. In addition, the contour attention branch is also deeply self-supervised with a penalized ground-truth contour regarding the output of the network. There are two different non-linear modules in our network: skip-attention block (Fig. \ref{fig:layers_skip}) and V-transition layer (Fig. \ref{fig:layers_v}). Each module is comprised of depth-wise separable convolutions, batch normalization \cite{ioffe2015batch}, rectified linear unit (ReLU) nonlinear activation function \cite{nair2010rectified}, skip connection, and channel-wise attention. Details of the architecture are described in the following subsections.

\subsection{Common Feature Extraction}
The skip-attention block (Fig. \ref{fig:layers_skip}) is first used to extract common features (i.e., shared features in the following layers). Subsequently, the features are fed to the liver-prior, context, and contour sub-networks (Fig. \ref{fig:network}). The skip-attention block is composed of non-linear transformation series: separable convolutions, batch normalization, and ReLU non-linear activation function (Fig. \ref{fig:layers_skip}). These transformations are skip connected for feature reuse. We introduced depth-wise separable convolutions \cite{chollet2017xception} in the skip-attention block rather than bottleneck \cite{szegedy2016rethinking} or compression \cite{huang2017densely} layers for more efficient use of parameters. The attention mechanism is applied to the final output to employ channel-wise attention, similar to \cite{hu2018squeeze}. Unlike \cite{hu2018squeeze}, for simplicity, we directly applied a trainable channel-wise attention vector that is multiplied for each channel.

\subsection{Liver-Prior Inference and Auto-Context Algorithm}
% architecture of auto-context
The base architecture of the proposed network is the auto-context algorithm. Instead of stacking deep neural layers, our proposed network uses multiple shallow stacks of layers (Fig. \ref{fig:network}). The liver-prior and context layers are composed of V-transition layers, which are small V-net-like modules that include down and up transitions together with skip connections (Fig. \ref{fig:layers_v}). The channel-wise attention is applied to the features in the lower resolution. The two identical shape transitions are used in the liver-prior block to subtract each output prediction at a higher level (blue dotted box in Fig. \ref{fig:network}). The output is deeply supervised with the ground-truth label image. The dual-passing architecture effectively learns mutually complementary features for the accurate inference of the liver posterior. The objective function for deep supervision of liver-prior can be defined as follows
\begin{equation}
    L_{p}=\mathcal{D}\big( (V_{p}^0(S(\bold{x}))-V_{p}^1(S(\bold{x}))), \bold{y_{dl}} \big),
\label{eq:loss_lp}
\end{equation}
where $\bold{x}, S, \bold{y_{dl}}$ denotes input image, skip-attention block, and the ground-truth liver label at down-scaled resolution, respectively. $\mathcal{D}$ denotes the soft dice loss \cite{milletari2016v} and $V_p^i$ indicates the $i^{th}$ V-transition in the liver-prior sub-network. Finally, the output feature map is concatenated to the context features (i.e., output of the context sub-network; $V_c(S(\bold{x}))$) and is passed through an auto-context sub-network ($V_a$) for the final refinement:
\begin{equation}
    L_f=\mathcal{D}\bigg( V_a\Big( \Big[ V_c(S(\bold{x})), R_p(S(\bold{x})), V_\mathcal{C}(S(\bold{x})) \Big] \Big), \bold{y_l} \bigg),
\label{eq:loss_lf}
\end{equation}
where $R_p$ denotes the residual output of the liver-prior subnetwork, $V_\mathcal{C}$ indicates the contour V-transition described in the subsequent subsection, and $\bold{y_l}$ denotes the ground-truth liver label.\par

% transition layer description.
The V-transition architecture is visualized in Fig. \ref{fig:layers_v}. The down-transition process down-samples the feature map by a factor of two for each dimension through $2^3$ convolutions with stride = 2. Conversely, an up-transition process restores the dimensions through a de-convolution (i.e., transposed convolution). We designed a skip connection and channel-wise attention in the lower dimension. By contracting and expanding paths, the V-transition layer can extract more multiscaled features (i.e., higher receptive field). A $1^3$ convolution is applied to the final concatenated features for further propagation. The number of output channels is as illustrated in Fig. \ref{fig:network}.

\subsection{Understanding the Network}
Let vectors \(\bold{x}=\{x_i\in\textit{R}, i\in\mathbb{R}^3\}\) and \(\bold{y}=\{y_i\in\{0,1\}, i\in\mathbb{R}^3\}\) represent the input image and ground-truth label, respectively. The objective of the given segmentation problem is to determine the optimal solution for modeling a conditional probability distribution, \(p(\bold{y}|\bold{x})\), by determining the maximum a posterior, as presented below:
\begin{equation}
    \bold{\theta}^*=\argmax_\bold{\theta}p(\bold{y}|\bold{x}; \bold{\theta})=\argmax_\bold{\theta}p(\bold{x}|\bold{y}; \bold{\theta})p(\bold{y}),
\end{equation}
\noindent
where \(\bold{\theta}\) is a parameter set for classifiers. However, it is significantly difficult to model the likelihood (i.e., \(p(\bold{x}|\bold{y}; \bold{\theta})\)) and prior (i.e., \(p(\bold{y})\)); moreover, solving the decomposed posterior with a generative approach easily yields inaccurate results, primarily owing to the difficulty in likelihood and prior estimations. Our proposed network iteratively solves the posterior directly using the auto-context method \cite{tu2010auto}. In auto-context, the previous classification map is used as a shape feature (i.e., the term ``context" is used in the original paper) for additional classification. Setting \(t\) as a discrete time value, the auto-context is formulated as
\begin{equation}
    p^{(t)}\Big(\bold{y}|\bold{x}, p^{(t-1)}(\bold{y}|\bold{x}; \bold{\theta}_{(t-1)}); \bold{\theta}_t\Big) \longrightarrow p(\bold{y}|\bold{x}; \bold{\theta}^*).
\end{equation}
\noindent
Unlike the previous approaches \cite{tu2010auto, salehi2017auto}, we combined the shape-feature extraction procedure with a single-passing neural network. The output of our proposed network for time \(t\) can be formulated as
\begin{equation}
    p^{(t)}\Big(\bold{y}|\bold{x}, \Tilde{p}^{(t)}(\bold{y}|\bold{x}; \bold{\theta}_t); \bold{\theta}_t\Big),
\end{equation}
\noindent
where \(\Tilde{p}\) is a probability map of shape-residual sub-network (i.e., the blue dotted box in Fig. \ref{fig:network}). Applying deep supervision (i.e., auxiliary classifiers), we could obtain a single-passing neural network embedded with a previous posterior. Thus, we avoided using separated classifiers and storing previous classification maps.

\subsection{Self-Supervising Contour Attention}
From the base architecture of the aforementioned auto-context framework, we extended our network with an explicit focus on contour features. The primary differences from the original work \cite{chung2018deeply} are the automatization of the training procedure and removal of the categorical classification loss. Unlike defining the threshold and manipulating it manually during the iteration, we employed a penalized contour soft loss with respect to the output predictions of the network. We first calculated the contour weighting map that has larger values for the misclassified contour as follows:
\begin{equation}
    \widehat{\Gamma_c}=\Gamma_c \otimes {\bold{\Tilde{y}_l}}^{-1},
\label{eq:closs3}
\end{equation}
where $\Gamma_c$, $\otimes$, and ${\bold{\Tilde{y}_l}}^{-1}$ indicate the ground-truth contour image, element-wise multiplication operator, and the final inverse liver prediction, respectively. The ground-truth contour image contains a value of 1 for the contour and 0 elsewhere. For the inverse prediction, we applied $\bold{\Tilde{y}}_{\bold{l}i} = 1-\bold{\Tilde{y}}_{\bold{l}i}$ for every $i^{th}$ voxel, where $\bold{\Tilde{y}_l}$ is the final output prediction of a foreground liver after softmax operation. Finally, we applied a penalized contour loss as follows:
% \begin{equation}
% \begin{split}
%     L_{c}=&\psi(\bold{\Tilde{y}_c}, \Gamma_c, \widehat{\Gamma_c})\\
%     &=-\sum_{i\in \Omega}{\big(w_0(1-\Gamma_c(i))\text{log}(1-\bold{\Tilde{y}_c}(i))+w_1\Gamma_c(i)\widehat{\Gamma_c}(i)\text{log}(\bold{\Tilde{y}_c}(i))\big)},
% \label{eq:loss}
% \end{split}
% \label{eq:closs4}
% \end{equation}

\begin{equation}
    L_{\mathcal{C}}=-\sum_{i\in \Omega}{\big(w_0(1-\Gamma_{c,i})\text{log}(1-\bold{\Tilde{y}}_{\bold{c},i})+w_1\Gamma_{c,i}\widehat{\Gamma_{c,i}}\text{log}(\bold{\Tilde{y}}_{\bold{c},i})\big)},
\label{eq:closs4}
\end{equation}
where $\bold{\Tilde{y}_c}$ is the output prediction of the contour after softmax operation, $w_c$ denotes class-specific weights for class $c$, and $\Omega$ indicates the dimensions of the image (i.e., $\Omega\in\mathbb{R}^3$). Consequently, the contour loss includes sparse contour attention based on the final output (\ref{eq:closs3}), which is employed to penalize the confident output of the network at each iteration. The difference between the proposed loss function and the focal loss \cite{lin2017focal} is that the proposed self-supervision is intended to penalize the confident output regarding the final liver prediction rather than the confidence of the contour itself.

\subsection{Learning the Network}
The task of the given learning system is to maximize the posterior, $p(\bold{y}|\bold{x})$. To effectively model the probability distribution, we attempted to train our network model to map the segmentation function \(\phi(\bold{x}):\bold{x}\longrightarrow\{0,1\}\) by minimizing the following loss function:
% \begin{equation}
% \begin{split}
%     \textit{L}(\bold{x}, \bold{y_l}, \Gamma_c;W)=&\mathcal{D}(\bold{\Tilde{y}_l}, \bold{y_l})+\alpha\mathcal{D}(\bold{\Tilde{y}_{dl}}^0-\bold{\Tilde{y}_{dl}}^1, \bold{y_{dl}})+\\
%     &\beta\chi(\bold{\Tilde{y}_c}, \Tilde{\Gamma}_c)+\gamma\|W\|_2^2,
% \label{eq:loss}
% \end{split}
% \end{equation}
\begin{equation}
    L=L_f+\alpha L_p+\beta L_\mathcal{C}+\gamma\|W\|_2^2,
\label{eq:loss}
\end{equation}
where $L_f$, $L_p$, and $L_\mathcal{C}$ indicate objective functions defined at the final output (\ref{eq:loss_lf}), shape prior (\ref{eq:loss_lp}), and contour (\ref{eq:closs4}) layers, respectively. $W$ is a whole set of network parameters. $\alpha$, $\beta$, and $\gamma$ are weighting parameters. The output of the network is obtained by applying softmax to the final output feature maps.\par

``Xavier" initialization \cite{glorot2010understanding} is used for initializing all the weights of the proposed network. While training the network, we fixed the loss parameters as $\alpha=\beta=1$ and $\gamma=0.1$ in (\ref{eq:loss}). We used the rectified Adam optimizer \cite{liu2019variance} with a batch size of 4 and learning rate of 0.001. We decayed the learning rate by multiplying 0.5 for every 10 epochs. We trained the network for 100 epochs using an Intel i9-7900X desktop system with 3.30 GHz processors, 128 GB of memory, and Nvidia Titan RTX (24 GB) GPU machine. We implemented the network using the PyTorch framework. It took 2h to complete all the training procedures.

% \footnote{The segmentation of the liver competition conducted at the 2007 Medical Image Computing and Computer Assisted Intervention conference.}
% \footnote{3D Image Reconstruction for Comparison of Algorithm Database (https://www.ircad.fr/research/3dircadb)}
% \footnote{The Combined Healthy Abdominal Organ Segmentation challenge (https://doi.org/10.5281/zenodo.3367758)}

\subsection{Data Preparation and Augmentation}
We acquired 180 subjects in total: 90 subjects from a publicly available dataset\footnote{https://doi.org/10.5281/zenodo.1169361} in \cite{gibson2018automatic}, 20 subjects from MICCAI-SLiver07 dataset \cite{heimann2009comparison}, 20 subjects from 3Dircadb\footnote{https://www.ircad.fr/research/3dircadb}, 20 subjects from CHAOS challenge\footnote{https://doi.org/10.5281/zenodo.3367758}, and additional 30 annotated subjects with the help of clinical experts in the field. In the dataset, the slice thickness ranged from $0.5-5.0$mm and pixel sizes ranged from $0.6-1.0$mm.\par

The whole dataset was separated into three sets: training, validation, and testing. We first randomly shuffled the dataset and separated 80 images for testing. The remaining 100 images were used for training based on a two-fold cross-validation (i.e., 50 training images and 50 validation images). We resampled all abdominal CT images into $256\times256\times64$. We pre-processed the image using fixed windowing values: level=10 and width=700 (i.e., we clipped the intensity values under $-340$ and over $360$). After re-scaling, we normalized the input images into the range [0-1] for each voxel. On-the-fly random affine deformations were subsequently applied to the dataset for each iteration with an 80\% probability.

\section{Experiments}
In our experiments, we evaluated the performance in terms of accuracy and generalization of our proposed network by comparing these results with those of the other state-of-the-art FCN-based models. We used 3D U-net \cite{cciccek20163d}, V-net \cite{milletari2016v}, deeply supervised network (DSN) \cite{dou20173d}, VoxResNet \cite{chen2017voxresnet}, DenseVNet \cite{gibson2018automatic}, AGU-net \cite{schlemper2019attention}, CENet \cite{chung2018deeply}, and our proposed network, AutoCENet for the performance evaluation.

\begin{table*}[t!]
\renewcommand{\arraystretch}{1.7}
\captionsetup{justification=centering, labelsep=newline}
\caption{Accuracy evaluation of the proposed network and the state-of-the-arts}
\label{table:sota}
\begin{tabularx}{\textwidth}{l||>{\centering\arraybackslash}X|>{\centering\arraybackslash}X|>{\centering\arraybackslash}X|>{\centering\arraybackslash}X|>{\centering\arraybackslash}X|>{\centering\arraybackslash}X}
\textbf{Methods} & \textbf{DSC} & \textbf{Precision} & \textbf{Sensitivity} & \textbf{HD [mm]} & \textbf{95\% HD [mm]} & \textbf{ASSD [mm]}\\
\hline
3D U-net \cite{cciccek20163d} & $0.95\pm0.01$ & $0.94\pm0.02$ & $0.96\pm0.02$ & $45.20\pm31.93$ & $7.77\pm12.71$ & $1.33\pm0.91$\\
V-net \cite{milletari2016v} & $0.95\pm0.02$ & $0.94\pm0.02$ & $0.95\pm0.03$ & $26.52\pm19.05$ & $5.38\pm3.94$ & $1.20\pm0.65$\\
DSN \cite{dou20173d} & $0.92\pm0.02$ & $0.88\pm0.04$ & $\textbf{0.97}\pmb{\pm}\textbf{0.01}$ & $28.63\pm23.85$ & $7.40\pm9.33$ & $1.77\pm1.05$\\
VoxResNet \cite{chen2017voxresnet} & $0.95\pm0.01$ & $\textbf{0.95}\pmb{\pm}\textbf{0.02}$ & $0.95\pm0.02$ & $18.67\pm11.15$ & $4.99\pm5.89$ & $1.11\pm0.49$\\
DenseVNet \cite{gibson2018automatic} & $0.83\pm0.05$ & $0.75\pm0.09$ & $0.94\pm0.03$ & $37.19\pm14.52$ & $16.54\pm8.47$ & $3.98\pm1.69$\\
AGU-net \cite{schlemper2019attention} & $0.95\pm0.01$ & $0.94\pm0.03$ & $0.96\pm0.01$ & $31.57\pm22.22$ & $8.56\pm13.52$ & $1.34\pm1.07$\\
CENet \cite{chung2018deeply} & $0.95\pm0.01$ & $0.95\pm0.02$ & $0.96\pm0.01$ & $16.68\pm8.87$ & $3.55\pm1.36$ & $0.94\pm0.38$\\
\textbf{AutoCENet} & $\textbf{0.96}\pmb{\pm}\textbf{0.01}$ & $\textbf{0.95}\pmb{\pm}\textbf{0.02}$ & $\textbf{0.97}\pmb{\pm}\textbf{0.01}$ & $\textbf{14.96}\pmb{\pm}\textbf{4.25}$ & $\textbf{2.92}\pmb{\pm}\textbf{1.12}$ & $\textbf{0.82}\pmb{\pm}\textbf{0.32}$\\
\end{tabularx}
\end{table*}

\subsection{Evaluation Metrics}
The segmentation results were evaluated using the F1 score, precision, sensitivity, Hausdorff distance (HD), and average symmetric surface distance (ASSD). The F1 score is defined as follows:

\begin{equation}
    F_1=2\times\frac{Precision\times Sensitivity}{Precision+Sensitivity}.
\label{eq:f1}
\end{equation}
\noindent
Precision and sensitivity are defined by $P=\frac{TP}{TP+FP}$ and $S=\frac{TP}{TP+FN}$, where TP, FN, and FP are the numbers of true positive, false negative, and false positive voxels, respectively. The F1 score is equivalent to the dice coefficient \cite{milletari2016v}. The surface distance metrics were evaluated on a surface basis: HD, 95\% HD \cite{chung2018deeply}, and ASSD \cite{heimann2009comparison}. We applied 95\% of voxels for HD to exclude the 5\% of outlying voxels. The 95\% HD is a better-generalized evaluation of distance because there exist ground-truth variations on a portal vein region. Two-fold cross-validation was used to obtain the quantitative results listed in Table \ref{table:sota}.\par

\begin{figure}[t]
    \centering
        \vfil
        \subfloat[DSC.]{\includegraphics[width=\linewidth]{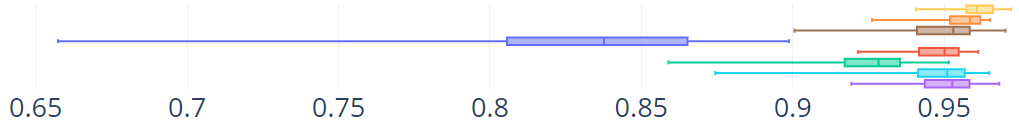}
        \label{fig:plot_sota_dsc}}
        \vfil
        \subfloat[HD in mm.]{\includegraphics[width=\linewidth]{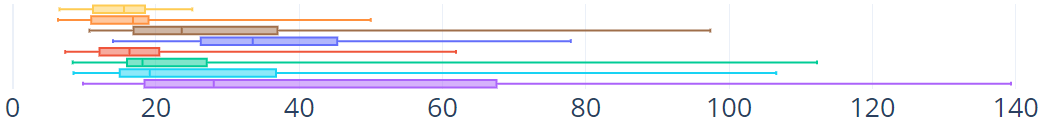}
        \label{fig:plot_sota_hd}}
        \vfil
        \subfloat[95\% HD in mm.]{\includegraphics[width=\linewidth]{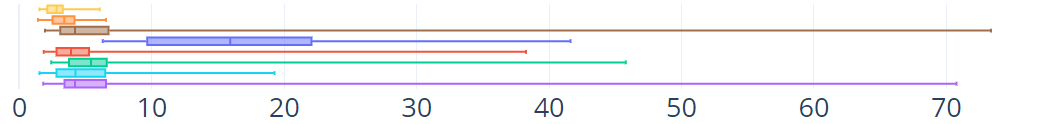}
        \label{fig:plot_sota_95hd}}
        \vfil
        \subfloat[ASSD in mm.]{\includegraphics[width=\linewidth]{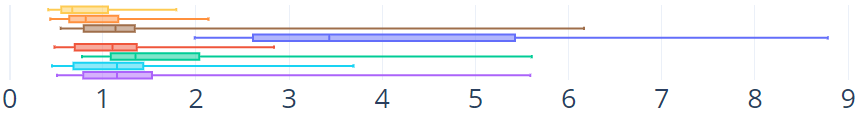}
        \label{fig:plot_sota_assd}}
        \vfil
        \subfloat[Sensitivity.]{\includegraphics[width=\linewidth]{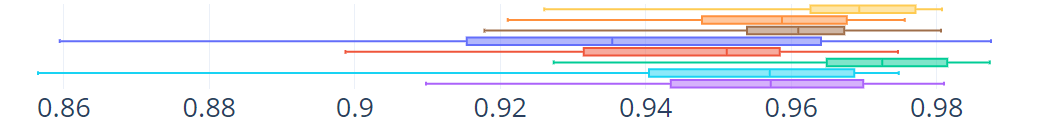}
        \label{fig:plot_sota_sens}}
        \vfil
        \subfloat[Precision.]{\includegraphics[width=\linewidth]{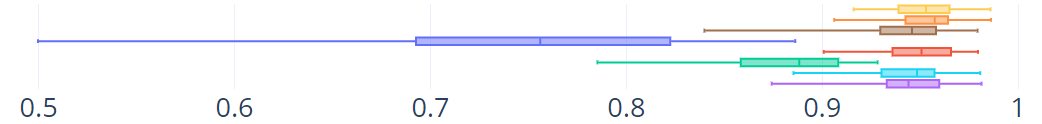}
        \label{fig:plot_sota_prec}}
        \vfil
        \subfloat{\includegraphics[width=\linewidth]{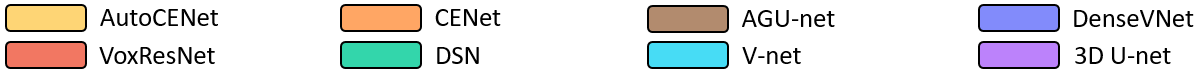}}
    \caption{Box plots of the evaluation metrics for state-of-the-art networks.}
    \label{fig:plot_sota}
\end{figure}

\subsection{Comparison}
Table \ref{table:sota} shows the quantitative results of liver segmentation. The results show that our proposed AutoCENet along with CENet \cite{chung2018deeply} outperformed other state-of-the-art networks;, moreover, our network showed better accuracy while using much fewer parameters than CENet. Table \ref{table:sota} lists that AutoCENet reduced ASSD by 12.77\% when compared to CENet. The lowest precision and sensitivity were presented by DenseVNet \cite{gibson2018automatic}. DenseVNet failed to segment the liver accurately because of two significant reasons: 1) the resolution of the network was too low and 2) shape prior was not robust. The excessively coarse dimensions of the network suffer from inaccurate segmentation in the original image resolution. Furthermore, $12^3$ resolution of shape prior is too small; moreover, the training images must be accurately and manually cropped for the robustness of the shape prior. There is no specific metric presented in the original paper \cite{gibson2018automatic} to crop the testing images automatically. The DSN \cite{dou20173d} showed high ASSD because the network was inferred from low resolution. The up-sampling process from $40\times 40\times 18$ demonstrated limitations in accurately delineating objects in the original resolution. The results indicate that multiple deep supervisions in DSN enforced the lower-level intermediate features to be discriminative, which resulted in degradation of the overall performance. The AGU-net also presented many false positives as opposed to the architectural design principle proposed in the original paper \cite{schlemper2019attention}. The spatial attention-gated units in AGU-net \cite{schlemper2019attention} failed to suppress irrelevant background regions as suggested. Conversely, VoxResNet \cite{chen2017voxresnet} showed the second minimum distance errors. The results of VoxResNet indicates that the auto-context algorithm successfully suppressed false positive responses. The box plots of the results listed in Table \ref{table:sota} are illustrated in Fig. \ref{fig:plot_sota}.

\begin{figure*}[t]
    \centering
        \vfil
        \subfloat[Ground-truth.]{\includegraphics[width=0.7in]{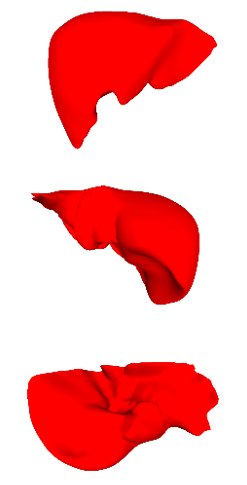}}
        \hfil
        \subfloat[3D U-net.]{\includegraphics[width=0.7in]{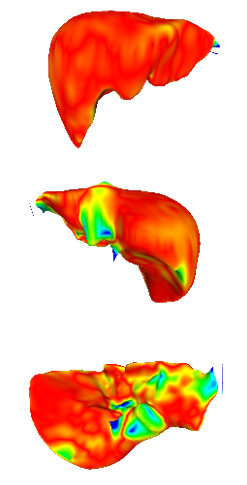}}
        \hfil
        \subfloat[V-net.]{\includegraphics[width=0.7in]{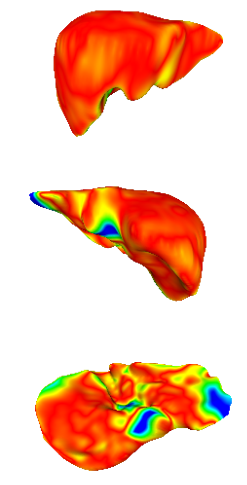}}
        \hfil
        \subfloat[DSN.]{\includegraphics[width=0.7in]{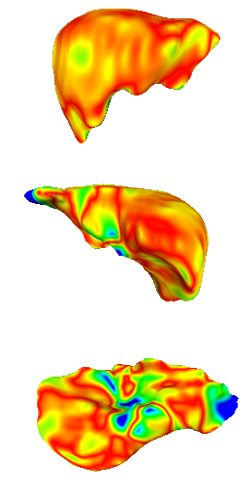}}
        \hfil
        \subfloat[VoxResNet.]{\includegraphics[width=0.7in]{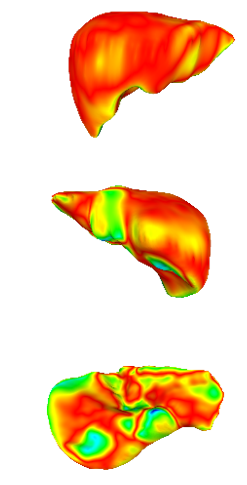}}
        \hfil
        \subfloat[DenseVNet.]{\includegraphics[width=0.7in]{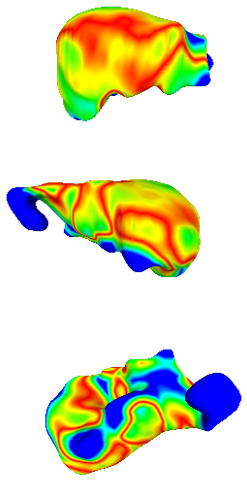}}
        \hfil
        \subfloat[AGU-net.]{\includegraphics[width=0.7in]{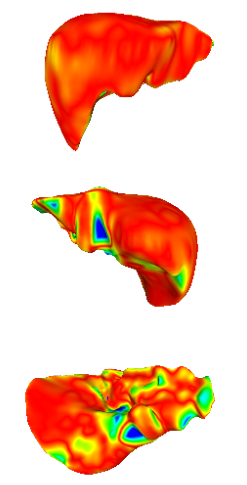}}
        \hfil
        \subfloat[CENet.]{\includegraphics[width=0.7in]{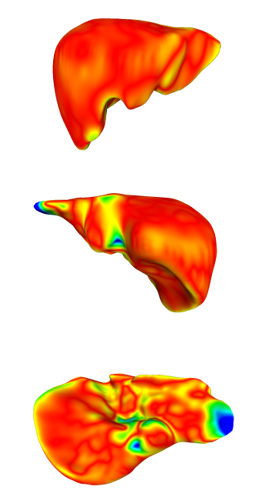}}
        \hfil
        \subfloat[AutoCENet.]{\includegraphics[width=0.7in]{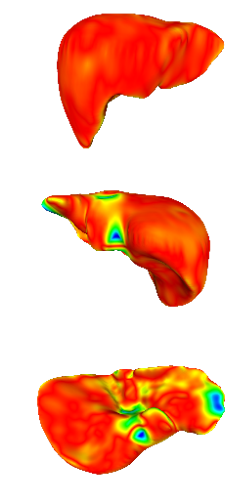}}\\[-1ex]
        % \vfil
        \subfloat{\includegraphics[width=3in]{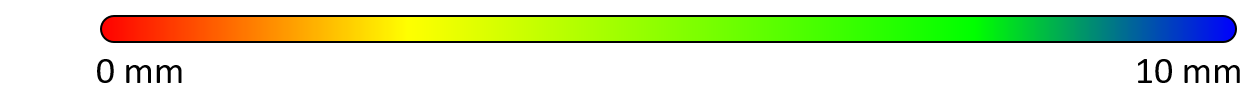}}
    \caption{Visualizations of the test results for state-of-the-art networks. The surface color is visualized based on the distance to the ground-truth surface.}
    \label{fig:vis}
\end{figure*}

\begin{table}[tb]
\renewcommand{\arraystretch}{1.8}
\captionsetup{justification=centering, labelsep=newline}
\caption{Performance of AutoCENet and its ablations}
\label{table:ablation}
\begin{tabularx}{\linewidth}{l|>{\centering\arraybackslash}X|>{\centering\arraybackslash}X|>{\centering\arraybackslash}X}
\centering
\textbf{Methods} & \textbf{DSC} & \textbf{HD [mm]} & \textbf{ASSD [mm]} \\
\hline
\textbf{AutoCENet} & $\textbf{0.96}\pmb{\pm}\textbf{0.01}$ & $\textbf{14.96}\pmb{\pm}\textbf{4.25}$ & $\textbf{0.82}\pmb{\pm}\textbf{0.32}$\\
\hline
% AutoNet+P & 0 & 0 & 0\\
AutoNet & $0.95\pm0.01$ & $20.18\pm8.79$ & $1.04\pm0.42$\\
AutoNet-att & $0.95\pm0.01$ & $25.73\pm17.06$ & $1.10\pm0.53$\\
AutoNet-A & $0.95\pm0.01$ & $33.25\pm22.80$ & $1.34\pm0.71$\\
AutoNet-R & $0.95\pm0.01$ & $37.99\pm25.09$ & $1.23\pm0.57$\\
AutoNet-AR & $0.94\pm0.01$ & $38.88\pm28.81$ & $1.32\pm0.61$\\
\hline
% Liver-Prior & $0.89\pm0.73$ & $3.72\pm0.31$ & $1.97\pm0.31$\\
% Liver-Prior-R & $0.89\pm0.73$ & $3.72\pm0.31$ & $1.97\pm0.31$\\
% \hline
% AutoCENet+P & $0.89\pm0.73$ & $3.72\pm0.31$ & $1.97\pm0.31$\\
AutoCENet+FC & $0.95\pm0.01$ & $27.56\pm20.34$ & $1.20\pm0.56$\\
% AutoCENet\_FCP & $0.89\pm0.73$ & $3.72\pm0.31$ & $1.97\pm0.31$\\
AutoCENet+MC & $0.95\pm0.01$ & $24.20\pm15.26$ & $1.14\pm0.56$\\
\end{tabularx}
\end{table}

\begin{figure}[t]
    \centering
    \subfloat[Input image.]{\includegraphics[width=1.4in]{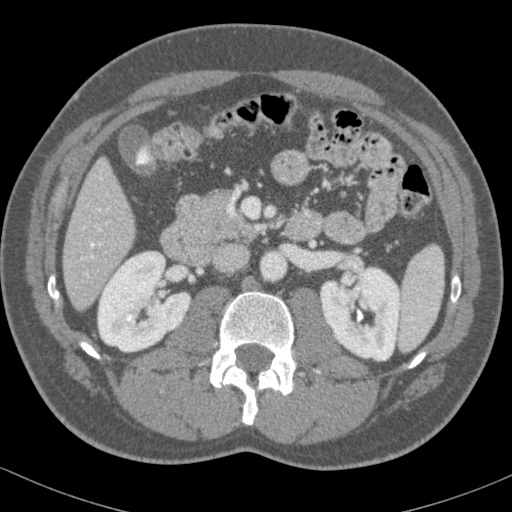}%
    \label{fig:vis_prior_image}}
    \hfil
    \subfloat[Ground-truth activation.]{\includegraphics[width=1.4in]{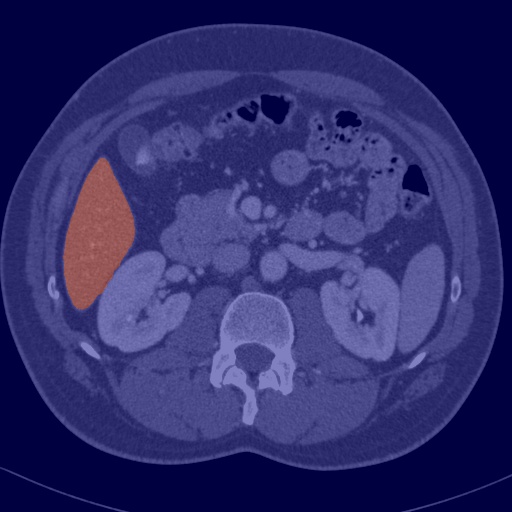}%
    \label{fig:vis_prior_gt}}\\[-1.5ex]
    % \vfil
    \subfloat[$\bold{\Tilde{y}_{dl}}^0$ in AutoNet.]{\includegraphics[width=1.4in]{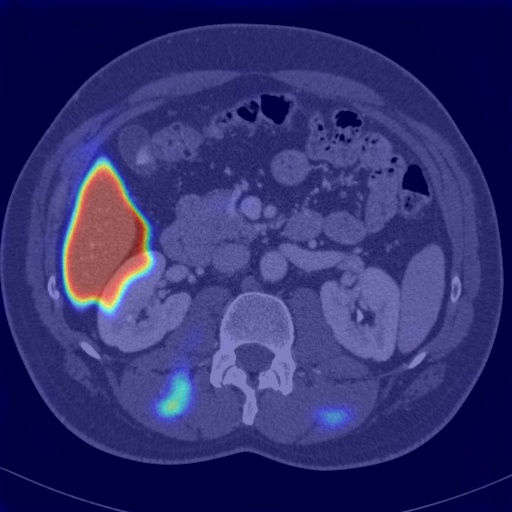}%
    \label{fig:vis_prior_s1}}
    \hfil
    \subfloat[$\bold{\Tilde{y}_{dl}}^1$ in AutoNet.]{\includegraphics[width=1.4in]{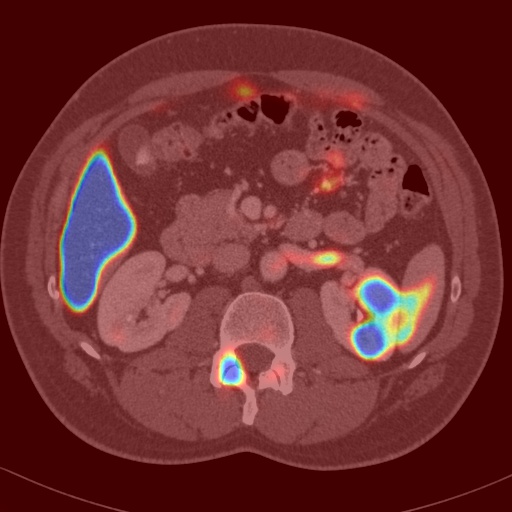}%
    \label{fig:vis_prior_s2}}\\[-1.5ex]
    % \vfil
    \subfloat[$\bold{\Tilde{y}_{dl}}^0-\bold{\Tilde{y}_{dl}}^1$ in AutoNet.]{\includegraphics[width=1.4in]{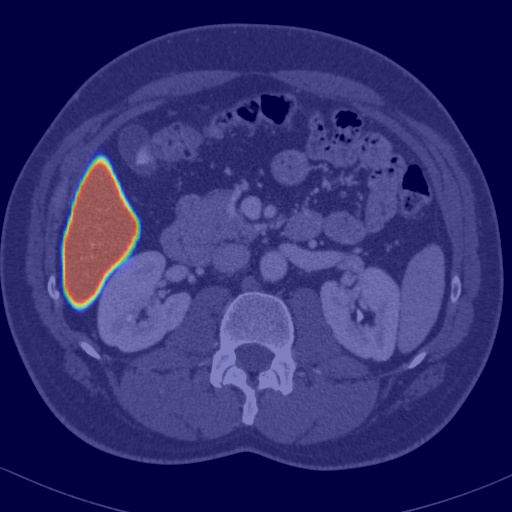}%
    \label{fig:vis_prior_s}}
    \hfil
    \subfloat[$\bold{\Tilde{y}_{dl}}$ in AutoNet-R.]{\includegraphics[width=1.4in]{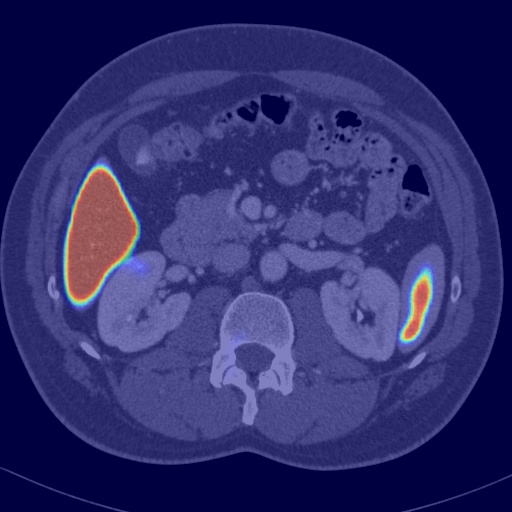}%
    \label{fig:vis_prior_r}}
    \vfil
    \includegraphics[width=\linewidth]{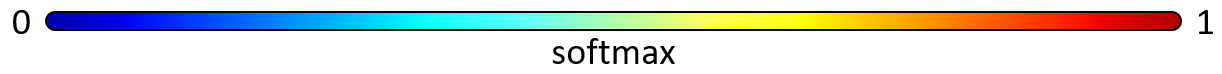}
    \caption{Liver prior estimations of AutoNet and AutoNet-R.}
    \label{fig:vis_prior}
\end{figure}

\subsection{Ablation Study}
We extended our experiments to verify the architectural components of the proposed network. We first validated the auto-context framework that does not exploit contour features (i.e., without contour loss, $L_c$ in (\ref{eq:loss}); AutoCENet). From the base auto-context framework, four additional ablations were studied: without channel-wise attention (AutoNet-att), without the auto-context part (i.e., AutoNet-A), without high-level residual inference (i.e., AutoNet-R), and without both auto-context and high-level residual inference (i.e., AutoNet-AR). In the case of AutoNet-A, we removed the deep supervision for the liver-prior network (Fig. \ref{fig:network}). For AutoNet-R, the high-level residual connection was modified to a sequential connection of V-transitions with the number of intermediate features equal to 48. AutoNet-AR employed both the modifications corresponding to AutoNet-A and AutoNet-R. The results (Table \ref{table:ablation}) showed that the accuracy of all the ablations was lower than the original AutoCENet. In AutoNet ablations, a significant increase pertaining to distance metrics were observed when the auto-context algorithm or residual shape prior were not employed. The results indicate that the auto-context framework and residual shape prior estimation jointly performed an important role in the final accuracy. The results of the liver-prior network with and without the residual inferences showed that the high-level residual connection boosted the performance of the liver-prior network. Sample visualizations of the liver priors are presented in the following subsection.\par

\begin{figure*}[t!]
    \centering
    \includegraphics[width=6.8in]{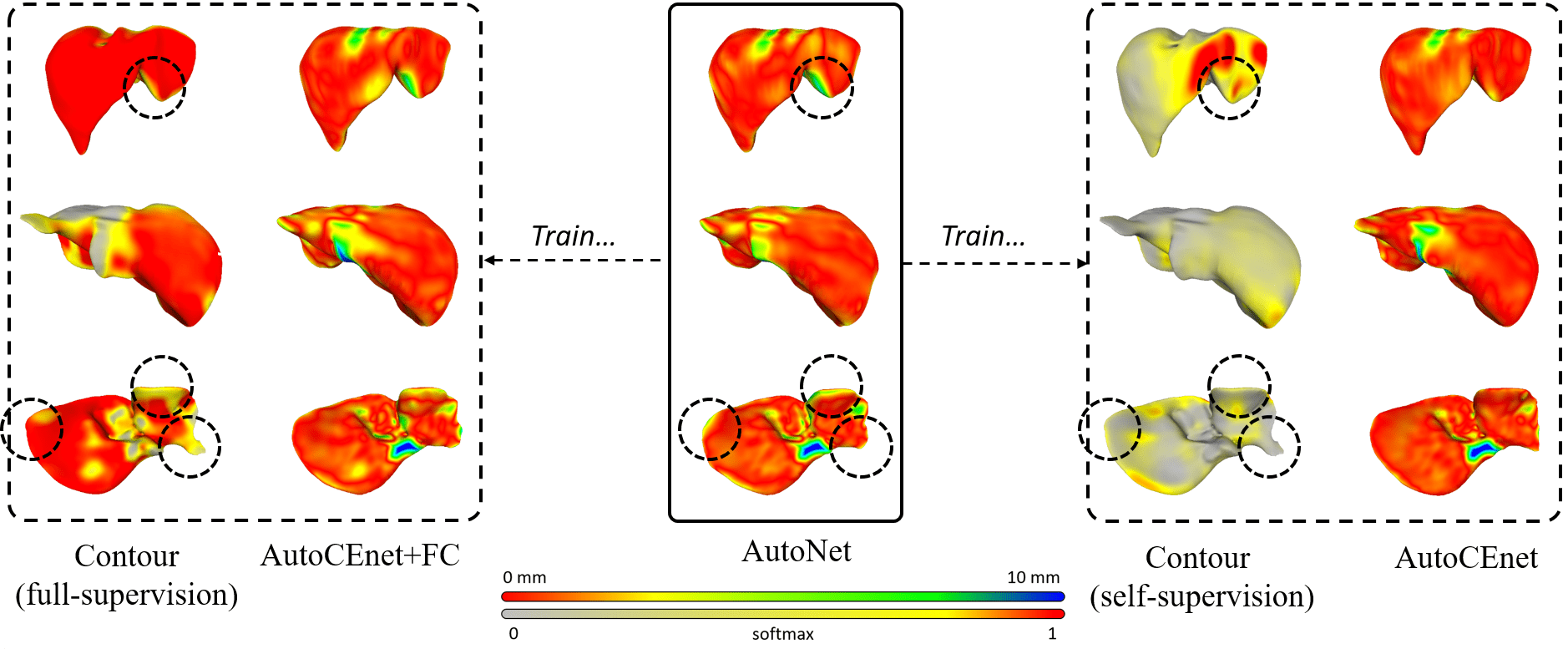}
    \caption{Visualizations of contour feature map and the final outputs after full training of AutoNet: full-contour supervision (left; AutoCENet+FC) and self-supervision (right; AutoCENet). The self-supervised contour feature map is sparser than that of the full supervision and is implicitly utilized as a strong contour attention. The ground-truth surface is used for visualizing the distribution of the contour feature. The softmax value is normalized into the range [0-1].}
    \label{fig:vis_contour_follow}
\end{figure*}

To verify the proposed self-supervised contour attention loss, we additionally experimented with two different contour losses: full-contour supervision of the ground-truth contour (AutoCENet+FC) and manual self-supervision, which was previously proposed in \cite{chung2018deeply}. The former full supervision was conducted without the penalization term presented in (\ref{eq:closs4}). The latter self-supervision was conducted by employing modified contour supervision, as presented in (\ref{eq:closs2}) \cite{chung2018deeply}. All the contour variants showed lower accuracy when compared to the original network. The performance of the AutoCENet+FC was more inferior than that of the AutoNet (Table \ref{table:ablation}) in terms of distance measures, indicating that enforcing the network to learn the full ground-truth contour image degrades the performance. Sample visualizations for the fully supervised and self-supervised contour feature maps are illustrated in the following subsection.\par

\subsection{Liver-Prior and Contour Feature Analysis}

Figure \ref{fig:vis_prior} shows the liver shape priors that were estimated with and without the proposed residual connection. The predicted probabilities clearly show the effectiveness of the high-level residual connection in shape prior estimation. The posterior of the liver from AutoNet-R (Fig. \ref{fig:vis_prior_r}) shows significant false positive responses when compared to the version with a residual connection (Fig. \ref{fig:vis_prior_s}). The two high-level predictions, i.e., Figs. \ref{fig:vis_prior_s1} and \ref{fig:vis_prior_s2}, were used as mutual complements to derive accurate liver prediction. The results indicate that the high-level residual inference shows an effective method to estimate accurate prior of a liver region without implementing a more complex and deep architecture of neural layers.\par

The contour feature map of a fully supervised network (i.e., using ground-truth contour supervision; AutoCENet+FC) was activated within overall contour regions (left box in Fig. \ref{fig:vis_contour_follow}). The figure illustrates that even with fully supervised training, the network failed to extract full-contour features accurately. Conversely, in the self-supervised network, the contour feature map was activated in the sparse regions (right box in Fig. \ref{fig:vis_contour_follow}). The sparse contour feature map acted as an implicit attention such that the network can concentrate more on the accurate delineation of boundary regions. By employing the self-supervised contour learning, the network demonstrated an improvement in the final segmentation. Figure \ref{fig:vis_contour_follow} illustrates the final output prediction of AutoNet and the following two networks: AutoCENet and AutoCENet+FC. The self-supervised contour responses did not correspond to the initial, weak contours from AutoNet (i.e., the initial sparse contour supervision starts from the weak parts of AutoNet results). A strong indication is that the self-supervised contour feature guides the network to better delineate object contours rather than learning the misclassified counterparts, as illustrated in Fig. \ref{fig:vis_contour_follow}. That is, the response of the contour feature successively changes pertaining to the current output prediction, which acts as implicit attention for the network. Note that the contour features are not complementary features that are to be merged for the final output prediction.\par

\begin{figure}[t!]
    \centering
    \subfloat[AutoCENet and state-of-the-art networks.]{\includegraphics[width=\linewidth]{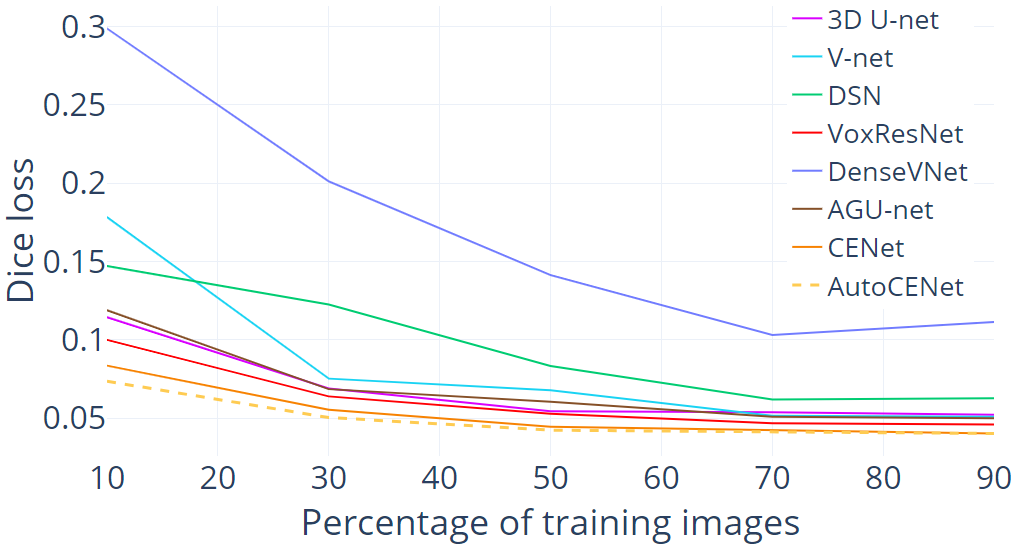}%
    \label{fig:plot_nfold_sota}}
    \vfil
    \subfloat[AutoCENet and its ablations.]{\includegraphics[width=\linewidth]{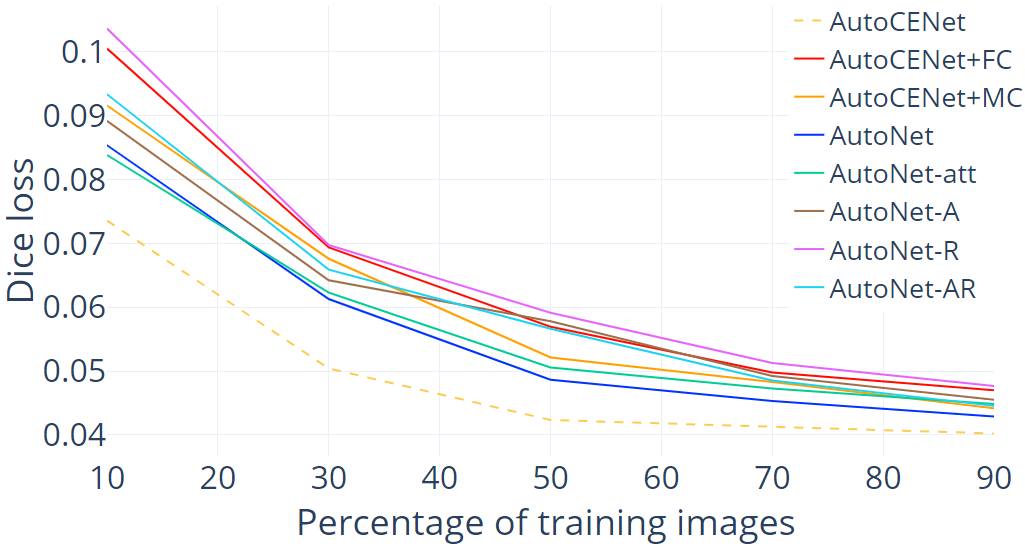}%
    \label{fig:plot_nfold_ablations}}
    % \vfil
    % \subfloat[Ground-truth activation.]{\includegraphics[width=\linewidth]{figures/plots/n_fold_cs.PNG}%
    % \label{fig:plot_nfold_cs}}
    \caption{N-fold cross-validation study of (a) the state-of-the-art networks and (b) the proposed network and its variants. The networks were cross validated by using 10\%, 30\%, 50\%, 70\%, and 90\% of the images for training out of a total of 100 images. The errors were calculated based on the 80 test images with dice loss.}
    \label{fig:plot_nfold}
\end{figure}

\subsection{Multiple N-Fold Validation}
Previous research has thoroughly investigated neural networks in an architectural perspective and verified their performances within individual metrics. However, limited academic research has been conducted to show the performance of generalization. To evaluate the performance of generalization, N-fold cross-validations were demonstrated for the presented networks. Figure \ref{fig:plot_nfold} illustrates the dice loss for the test images (i.e., 80 images) by training the network using 10\%, 30\%, 50\%, 70\%, and 90\% of training images out of the 100 images. The N-fold experiments approximately proxy the real-life deep learning problem and show an extremely generalized regularization analysis.\par

The overall test errors increased in a smaller percentage of training images. The proposed AutoCENet showed the best performance of generalization. AutoCENet did not over-fit the training images when compared to the other networks. The VoxResNet \cite{chen2017voxresnet} was the second-best out of other state-of-the-art networks. The fair performance of VoxResNet was obtained owing to its auto-context algorithm. The severe errors in DenseVNet \cite{gibson2018automatic} were caused by weak representative shape prior, as discussed in the aforementioned evaluations.\par

The ablation networks of AutoCENet showed comparable performances to the other state-of-the-art networks (Fig. \ref{fig:plot_nfold_ablations}. Among AutoNet variations, AutoNet-R was the worst-performing network indicating that residual shape prior estimation performs an important role in an auto-context algorithm. In the cases of contour variants, full supervision of contour (i.e., AutoCENet+FC) showed the worst performance (Fig. \ref{fig:plot_nfold_ablations}).

\section{Discussion}
In recent years, the employment of shape priors or neural networks has been the most promising method for the accurate segmentation of a liver. The proposed network avoided using the shape priors because the performance can be highly dependent on the trained shape variations. If the training set is insufficient, the algorithm easily breaks down owing to the quality of the trained prior. Our proposed auto-context algorithm introduced a high-level residual shape prior estimation process that robustly acquired the liver posterior. The embedded liver probability map acted as a post-inference prior, which can be further used for the final accurate classification in an auto-context framework. Consequently, a single-passing auto-context neural network was established without separate classification series, as presented in \cite{tu2010auto, salehi2017auto}. The primary underlying principle of the basic auto-context architecture is that the performance of generalization can be achieved by a robust estimation of the overall shape of a liver. In that perspective, high-level residual shape estimation in a lower resolution can successfully achieve the desired task. The architecture suggests that deepening or widening the neural network is not the only way for complex tasks. Stacking layers sequentially results in difficulty in using parameters effectively and further degrades the regularization of the network. The study of ablation for residual connection demonstrated that the proposed method, which was designed as a task-dependent curriculum, significantly outperformed a simple sequential architecture.\par

The attention mechanism has demonstrated increasing applicability as a dominant method for modern neural networks. However, the attention mechanism demonstrates a limitation as it is a data-driven algorithm, which indicates that the performance completely relies on the training data distribution. A simple adaptation of the attention mechanism cannot improve the baseline network without explicit guidance. The experimental results showed that the self-attention mechanism presented in AGU-net \cite{schlemper2019attention} did not show significant improvement when compared to the basic 3D U-net \cite{cciccek20163d}. It is significantly difficult to create a neural network that focuses more attention on certain features that are useful for the final output. In this study, a self-supervising contour delineation was applied to the intermediate layer that is intended to implicitly guide the network, rather than giving explicit attention, to focus more attention on weak boundary regions that the network has failed to accurately delineate. The self-supervising mechanism was successfully embedded in the network and it improved the final accuracy without any extra false positives.

% The overall architecture of the proposed neural network, which exploited an auto-context and the contour self-supervision, suggests that a performance of generalization can be obtained by human-designed curriculum which means a domain-specific knowledge is still required in the modern application of neural networks.

\section{Conclusion}
The accurate segmentation of a liver is still a challenging task. Although deep learning demonstrates increasing applications, the lack of annotated medical image data results in difficulty in successfully deploying CNNs in the clinics. Therefore, improving generalization performance is one of the most important tasks for utilizing CNN. In this study, a CNN for liver segmentation was proposed to minimize generalization errors based on the human-designed curriculum (i.e., auto-context). The proposed method minimized the error between training and test images more than any other modern neural networks. In addition, the contour scheme was successfully employed in the network by introducing a self-supervising metric. Instead of exploiting the entire ground-truth contour or self-attention, sparse contours were trained explicitly so that the network can focus on its failures. Based on the experimental results, it was identified that the proposed method performed a significant role in improving accuracy. The newly presented multiple N-fold cross-validation studies also demonstrated the practical applicability of the networks in actual clinics.

\ifCLASSOPTIONcaptionsoff
  \newpage
\fi

% references section
\bibliographystyle{IEEEtran}
\bibliography{MyBiB}

\end{document}